\documentclass[runningheads]{llncs}
\usepackage{graphicx}
\usepackage{tikz}
\usepackage{adjustbox}
\usepackage{tabularx}
\usepackage{graphicx}
\usepackage{booktabs,siunitx} 
\usepackage{textcomp}
\usepackage[T1]{fontenc}
\usepackage{comment}
\usepackage{amsmath} 
\usepackage{amssymb}
\usepackage{bbm}
\usepackage{amsmath}
\usepackage[pagebackref=true,breaklinks=true,colorlinks,bookmarks=false]{hyperref}
\usepackage{cleveref}
\usepackage{caption}
\usepackage{subcaption}
\usepackage{microtype}
\microtypesetup{expansion=false}
\usepackage{booktabs}
\crefname{section}{Sec.}{Sections}
\crefname{figure}{Fig.}{Figs.}
\crefname{table}{Tab.}{Tabs.}
\crefname{equation}{Eq.}{Eqs.}
\crefname{appsec}{Appendix}{Appendices}
\usepackage[T1]{fontenc}
\usepackage{graphicx}
\begin{document}
\title{Re-thinking Mammography Transfer Learning: The Dataset-Informed Transfer Learning (DITL) Framework for Breast Cancer Screening and Lesion Diagnosis}
\titlerunning{DITL Framework for Breast Cancer Screening and Lesion Diagnosis}
%
\author{Adarsh Bhandary Panambur\inst{1}\orcidID{0009-0004-8067-1150} \and
Siming Bayer\inst{1,2}\orcidID{0000-0003-2874-4805} \and
Andreas Maier\inst{1}\orcidID{0000-0002-9550-5284}}
%
\author{Adarsh Bhandary Panambur\inst{1}\orcidID{0009-0004-8067-1150} \and
Siming Bayer\inst{1,2}\orcidID{0000-0003-2874-4805} \and
Andreas Maier\inst{1}\orcidID{0000-0002-9550-5284}}
\authorrunning{A. Bhandary Panambur et al.}
\institute{Pattern Recognition Lab, Friedrich-Alexander-Universität Erlangen-Nürnberg, Erlangen, Germany 
\and Siemens Healthineers, Erlangen, Germany \\
\email{adarsh.bhandary.panambur@fau.de}}

\maketitle              
\begin{abstract}
Enhancing classification performance in mammography remains a persistent challenge across both small curated datasets and large-scale clinical cohorts. Conventional transfer learning approaches often neglect dataset-specific characteristics, while recent neighborhood-informed methods have been restricted to narrow tasks with rigid formulations, limiting their scalability to population-level datasets. To address these challenges, we propose the Dataset-Informed Transfer Learning (DITL) framework, which integrates dataset-derived difficulty signals with neighborhood-based triplet supervision in a unified objective. DITL introduces two adaptive components: (i) Adaptive Difficulty-Weighted Cross-Entropy (A-DWCE), which assigns per-sample weights based on $k$-nearest neighbor label purity in a self-supervised feature space, and (ii) Adaptive Neighborhood Representation Triplet (A-NR-Triplet), which enforces intra-class compactness and inter-class separation using a learnable margin. Unlike focal loss, DITL requires no hyperparameter tuning, removes heuristic weighting and fixed margins, and incurs negligible computational overhead, yielding a robust and scalable optimization strategy. On the large-scale VinDR-Mammo dataset, DITL achieves state-of-the-art performance for whole-image breast density classification, with significant improvements across accuracy, F1-score, and AUC ($p<0.0001$). These results underscore the framework’s ability to scale effectively to population-level screening tasks, where class imbalance and subtle imaging differences pose substantial challenges. Beyond large cohorts, DITL also delivers consistent, statistically significant gains on small ROI datasets ($p<0.0001$). By bridging small-scale lesion analysis with large-scale density estimation, DITL establishes a clinically relevant, scalable, and generalizable framework for mammography classification, spanning the full breast cancer screening-to-diagnosis spectrum.

\end{abstract}

\section{Introduction}
Breast cancer continues to represent a major global health concern, with incidence and mortality steadily rising over the past decades. Projections suggest that annual diagnoses may exceed 3.2 million cases worldwide, contributing to approximately 1.1 million deaths each year \cite{3454-00}. Digital mammography remains the gold standard imaging modality for large-scale screening programs, allowing radiologists to identify suspicious findings that may indicate either benign or malignant disease at an early stage. An important aspect of mammographic evaluation is the assessment of breast density, which quantifies the proportion of fibroglandular tissue relative to fatty tissue \cite{3454-01}. High breast density not only elevates the long-term risk of developing breast cancer but also reduces lesion appearance, thereby complicating reliable detection. Risk stratification is further refined using the Breast Imaging Reporting and Data System (BI-RADS), a standardized framework that categorizes findings from 0 (incomplete) to 6 (biopsy-confirmed malignancy) \cite{3454-01}. This system directly informs clinical management: low categories typically prompt routine follow-up, intermediate scores may lead to additional imaging, and BI-RADS 4 or higher often results in a biopsy recommendation. Importantly, population studies indicate that women in several Asian regions more frequently present with dense breast tissue, which reduces interpretability and contributes to geographic disparities in cancer detection \cite{nguyen2022novel}. With the increasing global prevalence of breast cancer, the higher rates of dense tissue in certain populations, and the growing demand for screening across large cohorts, radiologists face substantial workload and diagnostic complexity. In this setting, computer-aided diagnosis (CAD) systems driven by deep learning have emerged, promising \cite{diaz2024artificial}. These approaches, developed over the past two decades, offer scalable support for breast density estimation, BI-RADS categorization, and lesion diagnosis, ultimately enhancing the efficiency and consistency of mammography-based screening and diagnosis.

The availability of high-quality datasets is a major challenge in the field of medical imaging classification with deep learning (DL). Acquiring fully annotated datasets, such as those for mammography in cancer diagnosis, is costly and time-consuming. Consequently, smaller datasets are typically more prevalent in the medical imaging domain. Transfer learning has emerged as the predominant DL technique to address this challenge. We recently proposed the Difficulty-Weighted Neighborhood Representation (DWNR) framework \cite{panambur2024enhancing}, which demonstrated that combining difficulty-aware weighting with neighborhood-informed triplet supervision could improve downstream lesion classification of region of interest (ROI) images in a small dataset setting. In this work, we propose the Dataset-Informed Transfer Learning (DITL) framework, which generalizes our earlier work and scales it from small ROI studies to large population-level cohorts. At its core, DITL leverages self-supervised learning (SSL) features to extract intrinsic dataset characteristics through nearest-neighbor analysis. These characteristics are then incorporated into training via two adaptive objectives: (i) an Adaptive Difficulty-Weighted Cross-Entropy (A-DWCE), which assigns normalized per-sample weights based on $k$-nearest neighbor label purity, and (ii) an Adaptive Neighborhood Representation Triplet (A-NR-Triplet) loss, which contrasts nearest same-class mean features against farthest different-class mean features under a learnable margin. By jointly optimizing these objectives, DITL provides a stable and dataset-informed supervisory signal that penalizes misclassifications while adapting to feature distributions at scale. 

The main contributions of this work are as follows:  
1) The development of a simple DITL framework for downstream classification. It uses dataset properties identified through SSL to adjust for misclassifications. This downstream framework assigns each sample in the training dataset a difficulty level. It also pairs each sample with a nearest and farthest neighbor based on class label, integrating these elements into the training process.  
2) Introduction of A-DWCE and A-NR-Triplet, which replace heuristic weighting and fixed-margin triplet loss with normalized difficulty-based weighting and a learnable margin, respectively.  
3) Comprehensive evaluation across four mammography datasets covers five clinically relevant tasks. These include breast density and BI-RADS categorization on whole mammogram images using the VinDr-Mammo dataset\cite{vindr}, as well as malignancy classification on ROI lesion  images using CESM \cite{khaled2022categorized}, CBIS-DDSM Mass and Calcification datasets \cite{lee2017curated}. 4) Comparison of SSL backbones, including DINO\cite{caron2021emerging} (Vision Transformer) and SimCLR\cite{chen2020simple} (ResNet-18), demonstrates the generalizability of DITL across architectures.  

\begin{figure}[h]
\centering
\includegraphics[width=0.75\textwidth]{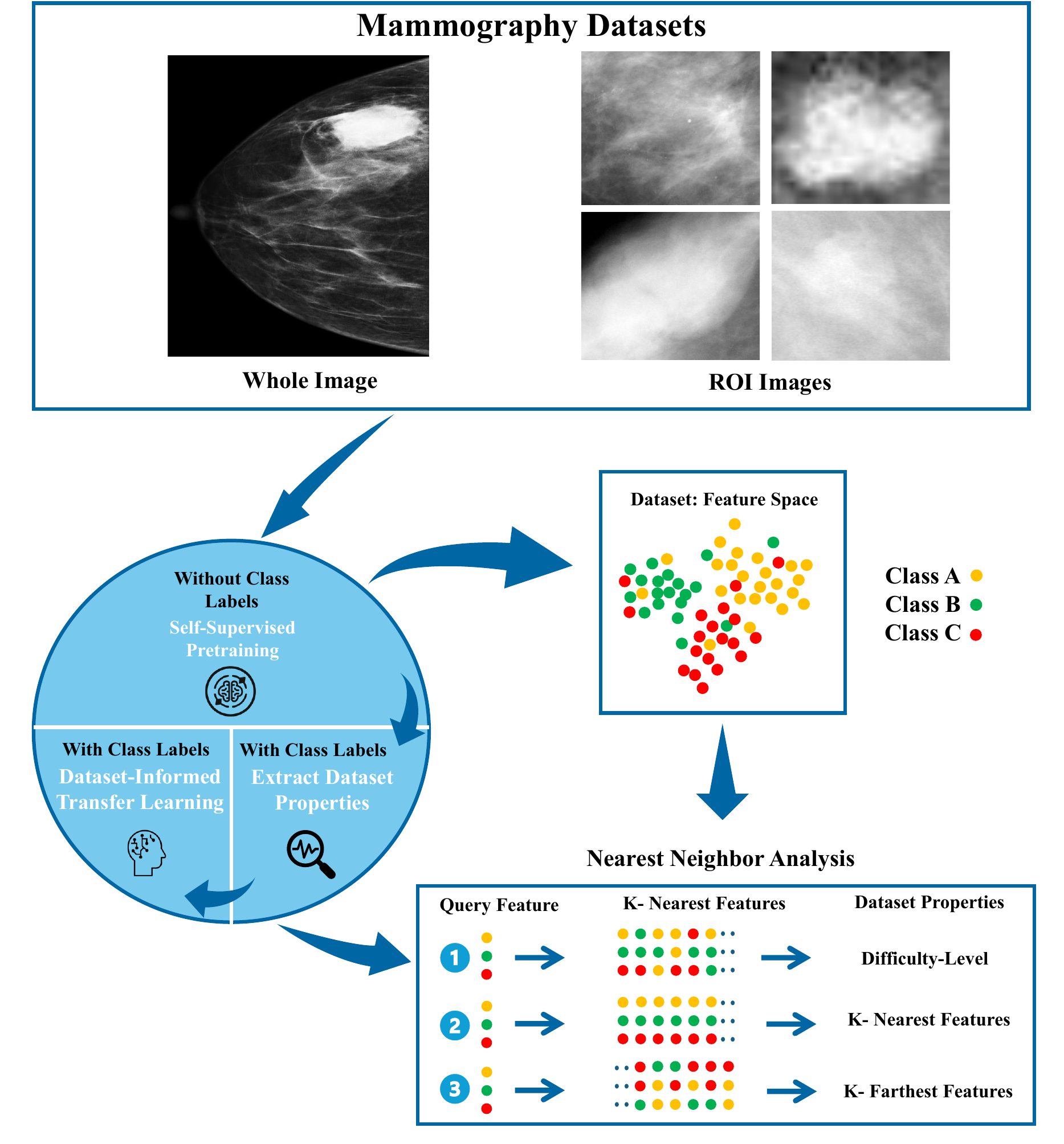}
\caption{Overview of the proposed 3-stage DITL downstream classification framework. 1) SSL pretraining without class labels; 2) Extracting dataset-specific features via nearest-neighbor analysis utilizing the SSL features and corresponding class labels; 3) Utilizing extracted properties for DITL.} 
\label{fig:fig1}
\end{figure}
Our methodology highlights the critical role of incorporating dataset-specific insights into the transfer learning process for mammography, demonstrating alignment across the full screening-to-diagnosis spectrum by improving performance not only in ROI-based lesion diagnosis but also in large-scale breast density estimation and BI-RADS risk assessment.

\section{Related Work}
In recent years, SSL has demonstrated its efficacy in deriving meaningful representations, which are instrumental for downstream tasks such as classification \cite{chen2020simple,caron2021emerging}. Numerous studies have also highlighted the advantages of employing pretraining with ImageNet and the concerned medical imaging datasets, and subsequent transfer learning to improve the efficacy and robustness of classification models \cite{azizi2021big}. Nonetheless, these pretraining strategies, even though they use many images without labels, still need large datasets to significantly improve classification performance. When trained with smaller datasets, these algorithms often struggle to learn the intrinsic properties and features, unique to the dataset distribution. This problem is further compounded because every dataset, influenced by its imaging modality or the particular distribution of abnormalities, exhibits unique intrinsic characteristics. Traditional transfer learning methods, which generally refine features initially learned from natural images, struggle to fully grasp these distinct aspects.  

Recent studies have explored various methodologies for extracting and applying dataset-specific characteristics, highlighting the crucial role of domain-specific knowledge in DL \cite{xie2021survey}. Tailored learning strategies, notably curriculum learning, have proven to be instrumental in augmenting the training process based on various levels of difficulty of samples and generalizability of deep learning algorithms, thereby elevating classification accuracy \cite{bengio2009curriculum}. The integration of additional clinical insights during the training phase has been demonstrated to enhance model robustness and operational efficiency. For instance, Nebbia et al. have illustrated the benefits of designing customized curricula based on extracted radiomic features to boost classification performance \cite{nebbia2021radiomics}. More recently, Panambur et al. introduced Attention-Guided Erasing (AGE), a data augmentation strategy that leverages attention head visualizations from DINO to weakly localize and erase background regions, thereby improving downstream transfer learning performance across both image- and patch-level mammography classification tasks \cite{panambur2025attention}. Additionally, the adoption of dataset-adaptive loss functions, such as focal loss, has been effective in directing the learning focus towards challenging samples and minority classes \cite{lin2017focal}.  

Building on these efforts, we recently proposed the Difficulty-Weighted Neighborhood Representation (DWNR) framework \cite{panambur2024enhancing}, which demonstrated that combining difficulty-aware weighting with neighborhood-informed triplet supervision could improve classification in CESM ROI analysis. While effective, DWNR suffered from several limitations: (i) the difficulty weights were defined heuristically and not normalized, leading to instability across datasets, (ii) the neighborhood representation triplet loss relied on a fixed margin, necessitating repeated margin experiments and restricting adaptability across varying feature distributions, and (iii) the number of neighbors $k$ had to be selected experimentally, which limited scalability to larger and more diverse cohorts. Furthermore, our prior evaluation was restricted to the CDD-CESM dataset \cite{khaled2022categorized}, constraining the generalizability of the framework beyond this specific modality and task. These limitations motivated the present DITL framework, which addresses these shortcomings by introducing adaptive weighting, a learnable triplet margin, and scalability across multiple datasets and tasks.

\section{Methods}

\cref{fig:fig1} shows an overview of the proposed DITL downstream classification pipeline. The framework consists of three stages. In the first stage, we perform self-supervised pretraining using established SSL models such as DINO \cite{caron2021emerging} and SimCLR \cite{chen2020simple}. Importantly, this pretraining is performed without using any class label information from the mammography datasets. In the second stage, we extract dataset-specific properties by applying nearest-neighbor analysis on the lower-dimensional feature representations obtained from the SSL backbone, combined with class label information. This stage yields difficulty scores as well as representative neighbor embeddings for each training sample. In the final stage, we incorporate these extracted properties into the transfer learning process by employing adaptive loss functions that guide downstream classification.

\subsection{Self-Supervised Pretraining}
We employ SSL backbones to obtain good representations of the images without any mammography labels. Specifically, we adopt DINO~\cite{caron2021emerging} with a ViT backbone and SimCLR~\cite{chen2020simple} with a ResNet backbone, both pretrained on ImageNet. DINO trains a student network to match the soft targets of an exponential moving average (EMA) teacher under multi-crop augmentations, thereby learning both global and local representations via a cross-entropy objective on softmax probabilities~\cite{caron2021emerging}. In contrast, SimCLR maximizes agreement between two augmented views of the same image and minimizes agreement across different images using the normalized temperature-scaled cross-entropy (NT-Xent) loss~\cite{chen2020simple}. For whole-image tasks on VinDR (breast density and BI-RADS), we use DINO only, as its multi-crop formulation captures global context while retaining local cues, which is essential for image-level screening assessments. For ROI-level classification (CESM, CBIS-DDSM Mass, CBIS-DDSM Calcification), we utilize both DINO and SimCLR to assess generalizability at the patch scale.  We extract embeddings from the final representation layer and use them as fixed features for the nearest-neighbor property extraction; the backbone is subsequently fine-tuned during DITL training.

\subsection{Extract Dataset Properties}

SSL pretraining methods such as DINO are well-known for learning robust lower-dimensional feature representations, as evidenced by their strong k-NN classification performance on ImageNet \cite{caron2021emerging}. This indicates that SSL backbones intrinsically capture sample distributions and can cluster semantically related images by leveraging both global and local features. Similarly, SimCLR achieves clustering of semantically similar samples by enforcing consistency across augmented views using a contrastive learning framework \cite{chen2020simple}.

Building on these representations, prior work has shown the benefits of incorporating dataset-derived insights into training objectives. For instance, Nebbia et al. \cite{nebbia2021radiomics} demonstrated the value of difficulty-aware weighting for curriculum learning where radiomics is utilized to formulate a learnable score, while Dwibedi et al. \cite{dwibedi2021little} proposed the use of nearest-neighbor features as positive pairs during pretraining to enrich semantic alignment. In our recently proposed DWNR framework \cite{panambur2024enhancing}, we also combined per-sample difficulty weighting with neighborhood-based supervision to improve mammography classification.

Inspired by these observations, we hypothesize that performing a nearest-neighbor analysis in the SSL feature space—while leveraging class labels—can serve as an effective mechanism to extract dataset-specific traits \cite{panambur2024enhancing}. However unlike DWNR \cite{panambur2024enhancing} where $k$ was chosen experimentally, in our work, for each dataset of size $N$, we select the number of neighbors $k$ according to the heuristic $k = \lceil \sqrt{N} \rceil$, which provides a scale-adaptive balance between locality and robustness across different dataset sizes. Using these neighborhoods, we compute three properties: 1) Per-sample difficulty levels, derived from the label purity of the $k$ nearest neighbors. A high purity indicates an easier sample, while low purity corresponds to harder cases located near decision boundaries; 2) $k$ nearest neighbor (KNN) feature, defined as the mean of the feature vectors of the closest neighbors belonging to the same class as input; 3) $k$  Farthest neighbor (KFN) features, defined as the mean of the most distant feature vectors belonging to different classes. These dataset properties are precomputed and stored for every sample, and are later incorporated into our adaptive loss functions during transfer learning.

\subsection{Dataset-Informed Transfer Learning (DITL)}

\subsubsection{Adaptive Difficulty-Weighted Cross-Entropy (A-DWCE)}
For each training sample, we estimate a difficulty score from the label purity of its $k$ nearest neighbors in the SSL feature space, where $k=\lceil\sqrt{N}\rceil$ and $N$ denotes the dataset size. Let $y_{query}$ be the class label of the query feature $f_{query}$ and $\hat{y}_{n}$ the label of its $n$-th nearest neighbor. The purity is
\[
\mathrm{purity} = \frac{1}{k}\sum_{n=1}^{k}\mathbb{1}\left(y_{query}=\hat{y}_{n}\right),
\]
and the difficulty is defined as,
\begin{equation}
w_{\mathrm{diff}} = 1 - \mathrm{purity}.
\label{eq:diff}
\end{equation}
To ensure stability, the per-sample $w_{\mathrm{diff}}$ values are normalized within each mini-batch. These weights are then applied to the per-sample cross-entropy loss, yielding the A-DWCE, mathematically shown as:
\begin{equation}
\mathcal{L}_{\mathrm{A\text{-}DWCE}} = - \sum_{i=1}^{N} w_{\mathrm{diff}}^{(i)} \, y_i \log \hat{y}_i ,
\label{eq:adwce}
\end{equation}
where $y_i$ denotes the ground-truth label and $\hat{y}_i$ the predicted probability for sample $i$. Although A-DWCE shares a conceptual resemblance with Focal loss \cite{lin2017focal} in down-weighting easy samples, it differs fundamentally in its weighting criterion. Instead of relying on prediction confidence, A-DWCE leverages nearest-neighbor label purity in the SSL feature space, which makes it more robust to noise and outliers and better aligned with dataset-specific difficulty. Moreover, unlike Focal loss, which requires extensive hyperparameter tuning of $\gamma$, A-DWCE is hyperparameter-free.

\subsubsection{Adaptive Neighborhood-Representation Triplet (A-NR-Triplet)}
In addition to difficulty weighting, we incorporate neighborhood structure via two dataset-derived representatives for each anchor: (i) the \emph{nearest same-class mean} ($\mathrm{KNN}_{+}$), computed as the mean of the $k$ closest neighbors sharing the anchor label , and (ii) the \emph{farthest different-class mean} ($\mathrm{KFN}_{-}$), computed as the mean of the $k$ most distant features from other classes \emph{measured from the anchor} in the SSL feature space. Here as well we use $k=\lceil\sqrt{N}\rceil$, where $N$ denotes the dataset size. Let $f_a$ denote the anchor embedding, and let $\mathrm{KNN}_{+}$ and $\mathrm{KFN}_{-}$ denote these mean vectors. We define the A-NR-Triplet as
\begin{equation}
\mathcal{L}_{\mathrm{A\text{-}NR\text{-}Triplet}} = \sum \max \!\left( d\!\left(f_a, \mathrm{KNN}_{+}\right) - d\!\left(f_a, \mathrm{KFN}_{-}\right) + m,\, 0 \right),
\label{eq:anr}
\end{equation}
where $d(\cdot)$ is the Euclidean distance and $m$ is a softplus-parameterized learnable margin. Using such margin allows DITL to adaptively scale inter-class separation during training, avoiding the fixed and heuristic margin in DWNR \cite{panambur2024enhancing} that often required manual tuning and reduced generalizability.

\subsection{Final Objective}
The final objective, termed DITL, integrates the two components directly:
\begin{equation}
\mathcal{L}_{\mathrm{DITL}} = \mathcal{L}_{\mathrm{A\text{-}DWCE}} + \mathcal{L}_{\mathrm{A\text{-}NR}\text{-}Triplet}.
\label{eq:ditl}
\end{equation}
This formulation ensures that misclassifications are penalized using normalized difficulty weights, while neighborhood structure enforces intra-class compactness and inter-class separation. This is used as a loss during downstream classifcation.

\begin{table}[t]
\centering
\caption{Class distributions for VinDR-Mammo. 
(a) Breast density classification (A--D). 
(b) BI-RADS categorization (1--5).}
\label{tab:vindr}
\small
\renewcommand{\arraystretch}{1.25}
\setlength{\tabcolsep}{8pt} 

\begin{tabular}{l|cccc|c}
\multicolumn{6}{c}{(a)} \\
\midrule
Split & \textbf{A} & \textbf{B} & \textbf{C} & \textbf{D} & \textbf{Total} \\
\midrule
Train & 64   & 1{,}314 & 10{,}412 & 1{,}808 & 13{,}598 \\
Valid & 16   & 214     & 1{,}818  & 352     & 2{,}400  \\
Test  & 20   & 380     & 3{,}060  & 540     & 4{,}000  \\
\bottomrule
\end{tabular}
\hspace{0.5cm}
\begin{tabular}{l|ccccc|c}
\multicolumn{7}{c}{(b)} \\
\midrule
Split & \textbf{1} & \textbf{2} & \textbf{3} & \textbf{4} & \textbf{5} & \textbf{Total} \\
\midrule
Train & 9{,}166 & 3{,}132 & 640 & 508 & 152 & 13{,}598 \\
Valid & 1{,}556 & 610     & 104 & 102 & 28  & 2{,}400  \\
Test  & 2{,}682 & 934     & 186 & 152 & 46  & 4{,}000  \\
\bottomrule
\end{tabular}
\end{table}

\begin{table}[t]
\centering
\caption{Class distributions for ROI-level datasets.}
\label{tab:roi}
\small
\renewcommand{\arraystretch}{1.25}
\setlength{\tabcolsep}{8pt} 
\begin{tabular}{l|l|ccc|c}
\toprule
\textbf{Dataset} & \textbf{Split} & \textbf{Class 1} & \textbf{Class 2} & \textbf{Class 3} & \textbf{Total} \\
\midrule
CESM          & Train & 309 & 193 & 243 & 745 \\
              & Valid &  61 &  34 &  35 & 130 \\
              & Test  &  46 &  23 &  47 & 116 \\
\midrule
Calcification     & Train & 389 & 438 & 446 & 1{,}273 \\
              & Valid &  85 &  90 &  98 &   273 \\
              & Test  &  67 & 130 & 129 &   326 \\
\midrule
Mass     & Train &  91 & 468 & 525 & 1{,}084 \\
              & Valid &  13 & 109 & 112 &   234 \\
              & Test  &  37 & 194 & 147 &   378 \\
\bottomrule
\end{tabular}
\end{table}

\section{Experimental Setup}
\subsection{Datasets}

We evaluate our framework on four mammography datasets spanning both whole-image screening tasks and ROI-level diagnostic tasks, with class distributions summarized in Tables~\ref{tab:vindr} and~\ref{tab:roi}. The VinDR-Mammo cohort is employed for two large-scale screening tasks: breast density classification (A--D) and BI-RADS categorization (1--5) \cite{vindr}. Both tasks exhibit substantial class imbalance, with the majority of cases corresponding to density C and BI-RADS 1 or 2, while categories D and BI-RADS 4/5 are comparatively under-represented. This reflects real-world screening distributions, where high-risk cases are less frequent yet clinically critical. The CDD-CESM dataset comprises 1003 contrast-enhanced spectral mammography images annotated into normal, benign, and malignant categories \cite{khaled2022categorized}. Compared to VinDR-Mammo, this dataset is considerably smaller and relatively balanced, though malignant cases are slightly over-represented. It serves as a controlled setting for ROI-level analysis. From the CBIS-DDSM dataset \cite{lee2017curated,clark2013cancer}, we include two ROI subsets focusing on calcification and mass abnormalities. Both subsets are moderately imbalanced: calcification cases contain relatively fewer benign-without-callback samples, whereas mass cases include a larger proportion of malignant findings relative to benign. These distributions highlight the heterogeneity and imbalance commonly encountered in diagnostic datasets. In summary, our evaluation spans (i) a large, clinically representative whole-image dataset with strong imbalance and (ii) smaller ROI-level datasets with varying degrees of imbalance, enabling a comprehensive assessment of DITL across both screening and diagnostic settings.

\subsection{Training Settings}

For DINO pretraining, we reused the implementation reported by Caron et al. \cite{caron2021emerging} and employed a ViT-S/16 model pretrained on ImageNet as the backbone. For SimCLR, we adopted the implementations of Chen et al. and Khosla et al. \cite{chen2020simple,khosla2020supervised}, using a ResNet-18 backbone together with dataset-specific augmentations from Panambur et al. \cite{panambur2022effect}. To accommodate single-GPU constraints and small dataset sizes, all pretrainings were run with a batch size of 64. Pretraining, dataset property extraction, and transfer learning were conducted separately for each dataset. We utilize the standard data preprocessing pipeline as reported in \cite{panambur2025attention,panambur2024enhancing}. For optimization, we used the Adam optimizer with a learning rate of $5\times10^{-6}$. Early stopping was applied, retaining the checkpoint with the lowest validation loss. 

\subsubsection{DITL Implementation Details}

For dataset property extraction, SSL features were computed for each training sample, and the number of neighbors was selected adaptively as $k = \lceil \sqrt{N} \rceil$, where $N$ denotes the dataset size. To reduce computational expense extensively, we perform the neighbor search using FAISS \cite{douze2024faiss,22faiss}. For each anchor sample, we stored the mean of its $k$ nearest same-class neighbors and the mean of its $k$ farthest different-class neighbors. Per-sample difficulty scores were derived from $k$-NN label purity, normalized within each mini-batch, and applied as weights in the cross-entropy loss. 

\subsubsection{Baselines and Evaluation}

In the transfer learning experiments, we compared standard cross-entropy (CE) loss and focal loss \cite{lin2017focal} to our proposed DITL objectives.  For focal loss, we evaluated a range of recommended $\gamma$ values \cite{lin2017focal} and report results from the best validation performance. Classification performance was assessed using accuracy, F1-score, and AUC. In order to ensure stability of the results, we run each experiments 5 times and report the mean and standard deviations.

\section{Results and Discussion}

\subsection{Whole-Image Classification}
Table~\ref{tab:results_vindr} summarizes the performance of different loss functions and state-of-the-art (SOTA) baselines on the VinDR-Mammo dataset for breast density (A--D) and BI-RADS (1--5) classification. These tasks represent clinically critical endpoints in population-scale mammography screening, where class imbalance and subtle imaging variations often hinder the performance of deep learning models.  

For breast density classification, DITL achieved the best overall results with an accuracy of 0.838, F1-score of 0.609, and AUC of 0.942. These gains are consistent across all metrics and surpass both conventional CE and Focal loss baselines. Compared to AGE \cite{panambur2025attention}, a recent SOTA method, DITL delivered a relative improvement of +1.8\% in F1 and +1.1\% in AUC, demonstrating the effectiveness of dataset-informed supervision in handling the severe imbalance between dense (C/D) and non-dense (A/B) categories. Importantly, the absolute F1 gains suggest that DITL improves recognition of under-represented dense cases, which are clinically the most challenging to classify.  For BI-RADS classification, both A-DWCE and DITL improved upon the CE and Focal loss baselines. A-DWCE obtained the highest accuracy (0.872), while DITL achieved the best F1 (0.585) and matched the best AUC (0.751). These results indicate complementary strengths: A-DWCE stabilizes training by normalizing sample difficulty, while DITL leverages neighborhood-informed triplet supervision to enhance discrimination of higher-risk BI-RADS categories. In practice, this translates to better differentiation between intermediate categories and more reliable detection of clinically suspicious cases. To verify robustness, we employed a two-tailed unpaired $t$-test to assess the statistical significance of classification performance differences across runs. The improvements achieved by DITL over baseline losses and SOTA methods were found to be highly significant ($p < 0.0001$).  

Overall, these results highlight three key insights. First, conventional adaptive losses such as Focal are insufficient for mammography-scale imbalance, where the rarity of high-density and high-BI-RADS cases demands dataset-specific supervision. Second, the proposed DITL framework consistently improves F1-scores, the most clinically relevant metric, demonstrating its ability to enhance minority-class recognition. Finally, the consistent and statistically significant improvements underscore the generality of DITL as a transferable framework for whole-image mammography screening tasks.

\begin{table*}[t]
\centering
\caption{Classification performance on the VinDR-Mammo test set. Metrics are reported as mean with standard deviation shown in parentheses. Best mean values are bolded. Asterisks indicate statistical significance.}
\label{tab:results_vindr}
\renewcommand{\arraystretch}{1.35}

\begin{minipage}{0.48\textwidth}
\centering
\subcaption{Breast density classification}
\begin{tabular}{l|ccc}
\toprule
\multicolumn{1}{c|}{\textbf{Method}} & \textbf{Acc.} & \textbf{F1} & \textbf{AUC} \\
\midrule
CE (baseline) &
  \begin{tabular}[c]{@{}c@{}}0.786\\ \scriptsize(0.029)\end{tabular} &
  \begin{tabular}[c]{@{}c@{}}0.576\\ \scriptsize(0.026)\end{tabular} &
  \begin{tabular}[c]{@{}c@{}}0.928\\ \scriptsize(0.008)\end{tabular} \\
\addlinespace
Focal ($\gamma{=}1.0$) &
  \begin{tabular}[c]{@{}c@{}}0.802\\ \scriptsize(0.011)\end{tabular} &
  \begin{tabular}[c]{@{}c@{}}0.584\\ \scriptsize(0.014)\end{tabular} &
  \begin{tabular}[c]{@{}c@{}}0.940\\ \scriptsize(0.001)\end{tabular} \\
\addlinespace
Nguyen et al.~\cite{nguyen2022novel} &
  \begin{tabular}[c]{@{}c@{}}--\\ --\end{tabular} &
  \begin{tabular}[c]{@{}c@{}}0.552\\ --\end{tabular} &
  \begin{tabular}[c]{@{}c@{}}--\\ --\end{tabular} \\
\addlinespace
AGE~\cite{panambur2025attention} &
  \begin{tabular}[c]{@{}c@{}}0.796\\ \scriptsize(0.023)\end{tabular} &
  \begin{tabular}[c]{@{}c@{}}0.591\\ \scriptsize(0.017)\end{tabular} &
  \begin{tabular}[c]{@{}c@{}}0.931\\ \scriptsize(0.001)\end{tabular} \\
\addlinespace
A\mbox{-}DWCE (Ours) &
  \begin{tabular}[c]{@{}c@{}}0.837\\ \scriptsize(0.004)\end{tabular} &
  \begin{tabular}[c]{@{}c@{}}0.596\\ \scriptsize(0.009)\end{tabular} &
  \begin{tabular}[c]{@{}c@{}}0.940\\ \scriptsize(0.003)\end{tabular} \\
\addlinespace
DITL (Ours) &
\begin{tabular}[c]{@{}c@{}}$\mathbf{0.838}^{\ast}$\\ {\scriptsize (0.005)}\end{tabular} &
\begin{tabular}[c]{@{}c@{}}$\mathbf{0.609}^{\ast}$\\ {\scriptsize (0.011)}\end{tabular} &
\begin{tabular}[c]{@{}c@{}}$\mathbf{0.942}^{\ast}$\\ {\scriptsize (0.003)}\end{tabular} \\

\bottomrule
\end{tabular}
\end{minipage}
\hfill
\begin{minipage}{0.48\textwidth}
\centering
\subcaption{BI-RADS classification}
\begin{tabular}{l|ccc}
\toprule
\multicolumn{1}{c|}{\textbf{Method}} & \textbf{Acc.} & \textbf{F1} & \textbf{AUC} \\
\midrule
CE (baseline) &
  \begin{tabular}[c]{@{}c@{}}0.851\\ \scriptsize(0.025)\end{tabular} &
  \begin{tabular}[c]{@{}c@{}}0.570\\ \scriptsize(0.012)\end{tabular} &
  \begin{tabular}[c]{@{}c@{}}0.742\\ \scriptsize(0.013)\end{tabular} \\
\addlinespace
Focal (best $\gamma$) &
  \begin{tabular}[c]{@{}c@{}}0.849\\ \scriptsize(0.013)\end{tabular} &
  \begin{tabular}[c]{@{}c@{}}0.572\\ \scriptsize(0.014)\end{tabular} &
  \begin{tabular}[c]{@{}c@{}}\textbf{0.751}\\ \scriptsize(0.010)\end{tabular} \\
\addlinespace
A\mbox{-}DWCE (Ours) &
  \begin{tabular}[c]{@{}c@{}}\textbf{0.872}\\ \scriptsize(0.009)\end{tabular} &
  \begin{tabular}[c]{@{}c@{}}0.569\\ \scriptsize(0.009)\end{tabular} &
  \begin{tabular}[c]{@{}c@{}}0.737\\ \scriptsize(0.015)\end{tabular} \\
\addlinespace
DITL (Ours) &
  \begin{tabular}[c]{@{}c@{}}$0.866$\\ {\scriptsize (0.030)}\end{tabular} &
\begin{tabular}[c]{@{}c@{}}$\mathbf{0.585}^{\!*}$\\ {\scriptsize (0.014)}\end{tabular} &
\begin{tabular}[c]{@{}c@{}}$\mathbf{0.751}$\\ {\scriptsize (0.014)}\end{tabular} \\

\bottomrule
\end{tabular}
\end{minipage}

\end{table*}

\begin{table}[t]
\caption{Performance comparison of DINO-based DITL with various loss functions. CE-Base (ImageNet-based ViT small) is the baseline. Highest values are bolded; asterisks indicate significant improvements of DITL-CE and DITL over the best-performing loss, p < 0.0001.}
\centering
\begin{tabularx}{\textwidth}{c| *{3}{>{\centering\arraybackslash}X} | *{3}{>{\centering\arraybackslash}X} | *{3}{>{\centering\arraybackslash}X}}
\toprule
& \multicolumn{3}{c|}{CESM} & \multicolumn{3}{c|}{Calcification} & \multicolumn{3}{c}{Mass} \\
\cmidrule{2-10}
Loss &  Acc. &  F1 &  AUC &  Acc. &  F1 &  AUC &  Acc. &  F1 &  AUC \\
\midrule
\begin{tabular}[c]{@{}c@{}}CE\\\scriptsize{(Base)}\end{tabular} & 
\begin{tabular}[c]{@{}c@{}}0.817\\\scriptsize{(0.014)}\end{tabular} & 
\begin{tabular}[c]{@{}c@{}}0.777\\\scriptsize{(0.022)}\end{tabular} & 
\begin{tabular}[c]{@{}c@{}}0.936\\\scriptsize{(0.003)}\end{tabular} & 
\begin{tabular}[c]{@{}c@{}}0.638\\\scriptsize{(0.018)}\end{tabular} & 
\begin{tabular}[c]{@{}c@{}}0.662\\\scriptsize{(0.020)}\end{tabular} & 
\begin{tabular}[c]{@{}c@{}}0.821\\\scriptsize{(0.017)}\end{tabular} & 
\begin{tabular}[c]{@{}c@{}}0.612\\\scriptsize{(0.054)}\end{tabular} & 
\begin{tabular}[c]{@{}c@{}}0.513\\\scriptsize{(0.033)}\end{tabular} & 
\begin{tabular}[c]{@{}c@{}}0.725\\\scriptsize{(0.013)}\end{tabular} \\
\addlinespace
\begin{tabular}[c]{@{}c@{}}CE\\\scriptsize{}\end{tabular} & 
\begin{tabular}[c]{@{}c@{}}0.838\\\scriptsize{(0.026)}\end{tabular} & 
\begin{tabular}[c]{@{}c@{}}0.799\\\scriptsize{(0.031)}\end{tabular} & 
\begin{tabular}[c]{@{}c@{}}0.931\\\scriptsize{(0.011)}\end{tabular} & 
\begin{tabular}[c]{@{}c@{}}0.658\\\scriptsize{(0.025)}\end{tabular} & 
\begin{tabular}[c]{@{}c@{}}0.685\\\scriptsize{(0.028)}\end{tabular} & 
\begin{tabular}[c]{@{}c@{}}0.841\\\scriptsize{(0.010)}\end{tabular} & 
\begin{tabular}[c]{@{}c@{}}0.641\\\scriptsize{(0.019)}\end{tabular} & 
\begin{tabular}[c]{@{}c@{}}0.561\\\scriptsize{(0.020)}\end{tabular} & 
\begin{tabular}[c]{@{}c@{}}0.741\\\scriptsize{(0.005)}\end{tabular} \\
\addlinespace
Focal & 
\begin{tabular}[c]{@{}c@{}}0.833\\\scriptsize{(0.030)}\end{tabular} & 
\begin{tabular}[c]{@{}c@{}}0.789\\\scriptsize{(0.031)}\end{tabular} & 
\begin{tabular}[c]{@{}c@{}}0.932\\\scriptsize{(0.004)}\end{tabular} & 
\begin{tabular}[c]{@{}c@{}}0.659\\\scriptsize{(0.016)}\end{tabular} & 
\begin{tabular}[c]{@{}c@{}}0.687\\\scriptsize{(0.016)}\end{tabular} & 
\begin{tabular}[c]{@{}c@{}}\textbf{0.841}\\\scriptsize{(0.009)}\end{tabular} & 
\begin{tabular}[c]{@{}c@{}}0.673\\\scriptsize{(0.011)}\end{tabular} & 
\begin{tabular}[c]{@{}c@{}}0.558\\\scriptsize{(0.009)}\end{tabular} & 
\begin{tabular}[c]{@{}c@{}}\textbf{0.771}\\\scriptsize{(0.009)}\end{tabular} \\
\addlinespace
\begin{tabular}[c]{@{}c@{}}A\mbox{-}DWCE\\\scriptsize{(Ours)}\end{tabular} & 
\begin{tabular}[c]{@{}c@{}}$0.839^{*}$\\\scriptsize{(0.009)}\end{tabular} & 
\begin{tabular}[c]{@{}c@{}}$0.806^{*}$\\\scriptsize{(0.015)}\end{tabular} & 
\begin{tabular}[c]{@{}c@{}}$0.939^{*}$\\\scriptsize{(0.003)}\end{tabular} & 
\begin{tabular}[c]{@{}c@{}}0.665\\\scriptsize{(0.008)}\end{tabular} & 
\begin{tabular}[c]{@{}c@{}}0.690\\\scriptsize{(0.004)}\end{tabular} & 
\begin{tabular}[c]{@{}c@{}}0.835\\\scriptsize{(0.002)}\end{tabular} & 
\begin{tabular}[c]{@{}c@{}}$\mathbf{0.677}^{*}$\\\scriptsize{(0.013)}\end{tabular} & 
\begin{tabular}[c]{@{}c@{}}{0.567}\\\scriptsize{(0.039)}\end{tabular} & 
\begin{tabular}[c]{@{}c@{}}0.752\\\scriptsize{(0.008)}\end{tabular} \\
\addlinespace
DITL & 
\begin{tabular}[c]{@{}c@{}}$\mathbf{0.860}^{*}$\\\scriptsize{(0.012)}\end{tabular} & 
\begin{tabular}[c]{@{}c@{}}$\mathbf{0.835}^{*}$\\\scriptsize{(0.020)}\end{tabular} & 
\begin{tabular}[c]{@{}c@{}}$\mathbf{0.943}^{*}$\\\scriptsize{(0.006)}\end{tabular} & 
\begin{tabular}[c]{@{}c@{}}$\mathbf{0.669}^{*}$\\\scriptsize{(0.017)}\end{tabular} & 
\begin{tabular}[c]{@{}c@{}}$\mathbf{0.693}^{*}$\\\scriptsize{(0.016)}\end{tabular} & 
\begin{tabular}[c]{@{}c@{}}0.837\\\scriptsize{(0.011)}\end{tabular} & 
\begin{tabular}[c]{@{}c@{}}0.677\\\scriptsize{(0.029)}\end{tabular} & 
\begin{tabular}[c]{@{}c@{}}$\mathbf{0.572}^{*}$\\\scriptsize{(0.011)}\end{tabular} & 
\begin{tabular}[c]{@{}c@{}}0.761\\\scriptsize{(0.006)}\end{tabular} \\
\bottomrule
\end{tabularx}

\label{tab1}
\end{table}

\subsection{ROI-Image Classification}

\cref{tab1} and \cref{tab2} present the classification performances of DITL using DINO and SimCLR backbones, respectively. The best results in each column are indicated in bold, and models with statistical significance at $p < 0.0001$ are marked with an asterisk. As seen in \cref{tab1}, the proposed DITL framework (DINO backbone) achieved the strongest performance improvements on the CESM dataset, with absolute gains of 2.2\%, 3.6\%, and 0.7\% in accuracy, F1, and AUC, respectively, compared to baseline methods (\cref{tab1}). In the calcification dataset, consistent gains are observed in accuracy and F1 when using DITL objectives, though the AUC of Focal loss remains slightly higher (\cref{tab1}). For mass classification, A-DWCE achieves superior accuracy compared to traditional losses, while Focal continues to yield the best AUC (\cref{tab1}).

Similarly, when applied with SimCLR features, DITL exhibits the same overall trends, with A-DWCE and the combined DITL objective outperforming conventional losses as seen in \cref{tab2}. Notably, DITL with SimCLR surpasses the curriculum learning approach of Nebbia et al. \cite{nebbia2021radiomics} in mass classification by almost 0.5\% absolute gain when compared with their reported ResNet-18 results, underscoring the benefits of dataset-informed weighting over radiomics-based curriculum learning, which also requires pixel-level labels.

Although the absolute gains of DITL over standard losses may appear modest in some cases, they remain statistically significant and clinically relevant, particularly in small ROI datasets. For example, improvements in F1-score reach up to 2.9\% in CESM and 2.0\% in mass classification, as confirmed by two-tailed unpaired $t$-tests ($p<0.0001$). Importantly, these dataset-informed benefits are obtained without additional computational overhead, since the nearest-neighbor analysis and extraction of dataset-specific properties are performed once offline after SSL pretraining. This design ensures that DITL training itself remains efficient and scalable while still leveraging dataset-specific difficulty levels and neighborhood features.

\subsection{General Observations}

Across both ROI-level and whole-image tasks, DITL consistently improved F1-scores, underscoring its strength in recognizing minority classes that are clinically the most challenging. These improvements were statistically significant ($p<0.0001$) and clinically meaningful, particularly for differentiating dense breast categories and intermediate BI-RADS scores. The complementary effects of A-DWCE and DITL highlight how difficulty normalization stabilizes training, while neighborhood-informed supervision enhances separation of higher-risk categories. Importantly, these improvements were consistent across SSL backbones, demonstrating the generality of DITL from small ROI datasets to large-scale screening cohorts.

\begin{table}[t]
\caption{Performance comparison of SimCLR-based DITL with various loss functions. CE-Base (ImageNet-based ViT small) is the baseline. Highest values are bolded; asterisks indicate significant improvements of DITL-CE and DITL over the best-performing loss, p < 0.0001}
\centering
\begin{tabularx}{\textwidth}{c| *{3}{>{\centering\arraybackslash}X} | *{3}{>{\centering\arraybackslash}X} | *{3}{>{\centering\arraybackslash}X}}
\toprule
& \multicolumn{3}{c|}{CESM} & \multicolumn{3}{c|}{Calcification} & \multicolumn{3}{c}{Mass} \\
\cmidrule{2-10}
Loss &  Acc. &  F1 &  AUC &  Acc. &  F1 &  AUC &  Acc. &  F1 &  AUC \\
\midrule
\begin{tabular}[c]{@{}c@{}}CE\\\scriptsize{(Base)}\end{tabular} & 
\begin{tabular}[c]{@{}c@{}}0.838\\\scriptsize{(0.016)}\end{tabular} & 
\begin{tabular}[c]{@{}c@{}}0.813\\\scriptsize{(0.022)}\end{tabular} & 
\begin{tabular}[c]{@{}c@{}}0.925\\\scriptsize{(0.005)}\end{tabular} & 
\begin{tabular}[c]{@{}c@{}}0.639\\\scriptsize{(0.0067)}\end{tabular} & 
\begin{tabular}[c]{@{}c@{}}0.667\\\scriptsize{(0.0062)}\end{tabular} & 
\begin{tabular}[c]{@{}c@{}}0.808\\\scriptsize{(0.0068)}\end{tabular} & 
\begin{tabular}[c]{@{}c@{}}0.601\\\scriptsize{(0.008)}\end{tabular} & 
\begin{tabular}[c]{@{}c@{}}0.522\\\scriptsize{(0.0054)}\end{tabular} & 
\begin{tabular}[c]{@{}c@{}}0.704\\\scriptsize{(0.0152)}\end{tabular} \\
\addlinespace
CE & 
\begin{tabular}[c]{@{}c@{}}0.845\\\scriptsize{(0.006)}\end{tabular} & 
\begin{tabular}[c]{@{}c@{}}0.816\\\scriptsize{(0.005)}\end{tabular} & 
\begin{tabular}[c]{@{}c@{}}0.928\\\scriptsize{(0.006)}\end{tabular} & 
\begin{tabular}[c]{@{}c@{}}0.653\\\scriptsize{(0.0095)}\end{tabular} & 
\begin{tabular}[c]{@{}c@{}}0.676\\\scriptsize{(0.009)}\end{tabular} & 
\begin{tabular}[c]{@{}c@{}}0.809\\\scriptsize{(0.0042)}\end{tabular} & 
\begin{tabular}[c]{@{}c@{}}0.643\\\scriptsize{(0.012)}\end{tabular} & 
\begin{tabular}[c]{@{}c@{}}0.528\\\scriptsize{(0.0229)}\end{tabular} & 
\begin{tabular}[c]{@{}c@{}}0.727\\\scriptsize{(0.0091)}\end{tabular} \\
\addlinespace
Focal & 
\begin{tabular}[c]{@{}c@{}}0.839\\\scriptsize{(0.013)}\end{tabular} & 
\begin{tabular}[c]{@{}c@{}}0.808\\\scriptsize{(0.018)}\end{tabular} & 
\begin{tabular}[c]{@{}c@{}}0.932\\\scriptsize{(0.002)}\end{tabular} & 
\begin{tabular}[c]{@{}c@{}}0.657\\\scriptsize{(0.022)}\end{tabular} & 
\begin{tabular}[c]{@{}c@{}}0.684\\\scriptsize{(0.019)}\end{tabular} & 
\begin{tabular}[c]{@{}c@{}}0.813\\\scriptsize{(0.009)}\end{tabular} & 
\begin{tabular}[c]{@{}c@{}}0.655\\\scriptsize{(0.015)}\end{tabular} & 
\begin{tabular}[c]{@{}c@{}}0.540\\\scriptsize{(0.023)}\end{tabular} & 
\begin{tabular}[c]{@{}c@{}}0.742\\\scriptsize{(0.009)}\end{tabular} \\
\addlinespace
\begin{tabular}[c]{@{}c@{}}A\mbox{-}DWCE\\\scriptsize{(Ours)}\end{tabular} & 
\begin{tabular}[c]{@{}c@{}}$0.853^{*}$\\\scriptsize{(0.006)}\end{tabular} & 
\begin{tabular}[c]{@{}c@{}}$0.823^{*}$\\\scriptsize{(0.005)}\end{tabular} & 
\begin{tabular}[c]{@{}c@{}}$0.933^{*}$\\\scriptsize{(0.001)}\end{tabular} & 
\begin{tabular}[c]{@{}c@{}}0.645\\\scriptsize{(0.011)}\end{tabular} & 
\begin{tabular}[c]{@{}c@{}}0.664\\\scriptsize{(0.014)}\end{tabular} & 
\begin{tabular}[c]{@{}c@{}}\textbf{0.813}\\\scriptsize{(0.004)}\end{tabular} & 
\begin{tabular}[c]{@{}c@{}}0.658\\\scriptsize{(0.009)}\end{tabular} & 
\begin{tabular}[c]{@{}c@{}}0.541\\\scriptsize{(0.012)}\end{tabular} & 
\begin{tabular}[c]{@{}c@{}}0.735\\\scriptsize{(0.009)}\end{tabular} \\
\addlinespace

\begin{tabular}[c]{@{}c@{}}DITL\\\scriptsize{(Ours)}\end{tabular} & 
\begin{tabular}[c]{@{}c@{}}$\mathbf{0.864}^{*}$\\\scriptsize{(0.005)}\end{tabular} & 
\begin{tabular}[c]{@{}c@{}}$\mathbf{0.842}^{*}$\\\scriptsize{(0.009)}\end{tabular} & 
\begin{tabular}[c]{@{}c@{}}$\mathbf{0.934}^{*}$\\\scriptsize{(0.002)}\end{tabular} & 
\begin{tabular}[c]{@{}c@{}}$\mathbf{0.662}^{*}$\\\scriptsize{(0.007)}\end{tabular} & 
\begin{tabular}[c]{@{}c@{}}$\mathbf{0.691}^{*}$\\\scriptsize{(0.017)}\end{tabular} & 
\begin{tabular}[c]{@{}c@{}}0.8120\\\scriptsize{(0.004)}\end{tabular} & 
\begin{tabular}[c]{@{}c@{}}$\mathbf{0.664}^{*}$\\\scriptsize{(0.014)}\end{tabular} & 
\begin{tabular}[c]{@{}c@{}}\textbf{0.542}\\\scriptsize{(0.025)}\end{tabular} & 
\begin{tabular}[c]{@{}c@{}}$\mathbf{0.749}^{*}$\\\scriptsize{(0.010)}\end{tabular} \\
\bottomrule
\end{tabularx}
\label{tab2}
\end{table}

\section{Conclusion}
In this work, we introduced the DITL framework, a unified strategy that integrates dataset-derived difficulty signals with neighborhood-informed supervision to improve mammography classification across both ROI- and whole-image tasks. By jointly optimizing A-DWCE and A-NR-Triplet losses, DITL removes heuristic weighting and fixed-margin constraints while maintaining efficiency through offline property extraction. Our experiments across four benchmark mammography cohorts consistently demonstrated that DITL outperforms conventional cross-entropy, focal loss, and some of the recent state-of-the-art methods. Importantly, the improvements are not only consistent but also statistically significant ($p<0.0001$) across multiple datasets. On small ROI datasets such as CESM and CBIS-DDSM, DITL achieves robust gains in accuracy and F1-score, validating its effectiveness in limited-data scenarios where conventional methods often struggle. On the large-scale VinDR-Mammo cohort, DITL scales effectively to clinically critical endpoints—breast density and BI-RADS categorization—where improvements in F1-scores highlight enhanced recognition of under-represented but high-risk categories. These results underscore the clinical relevance of DITL: improving classification of dense breasts and higher BI-RADS cases translates to better risk stratification and more reliable triage in real-world screening workflows. Beyond its performance gains, DITL offers notable practical advantages. In contrast to focal loss, it does not require additional hyperparameter tuning, thereby reducing the complexity of model optimization. Furthermore, since dataset properties are computed in an offline stage, the framework introduces negligible computational overhead during training. By directly leveraging cohort-specific distributions through dataset-informed weighting, DITL ensures both robustness across diverse settings and broad generalizability to heterogeneous clinical cohorts.

For future work, we aim to extend DITL to other algorithmic paradigms, including contrastive language–image pretraining approaches, where dataset-specific supervision could complement large-scale multimodal representations. A natural extension is to evaluate DITL beyond mammography, for instance in modalities such as chest X-rays, to further test generalizability. Additionally, integrating DITL within curriculum learning strategies is a promising direction, as prior work has shown that difficulty-aware curricula can substantially enhance medical imaging classification \cite{bengio2009curriculum,nebbia2021radiomics}. Such integration could yield even more adaptive training pipelines, combining global curriculum progression with dataset-specific weighting.

\begin{credits}
\subsubsection{\ackname} We express our gratitude to the authors and institutions behind the VinDR-Mammo, CBIS-DDSM and CDD-CESM dataset, as well as to TCIA \cite{lee2017curated,khaled2022categorized,clark2013cancer}. 
\subsubsection{\discintname}
The authors have no competing interests to declare that are relevant to the content of this article. 
\end{credits}

\bibliographystyle{splncs04}
\bibliography{mybibliography}
\end{document}